\documentclass{article}
\usepackage{microtype}
\usepackage{graphicx}
\usepackage{subfigure}
\usepackage{booktabs} 
\usepackage{multirow}
\graphicspath{{figs/}}
\usepackage{amsmath}
\usepackage{hyperref}
\usepackage[accepted]{icml2019}

\begin{document}

\twocolumn[
\icmltitle{AVA: A Financial Service Chatbot based on Deep Bidirectional Transformers}
\icmlsetsymbol{equal}{*}
\begin{icmlauthorlist}
\icmlauthor{Shi Yu}{to}
\icmlauthor{Yuxin Chen}{to}
\icmlauthor{Hussain Zaidi}{to}
\end{icmlauthorlist}
\icmlaffiliation{to}{The Vanguard Group, Malvern, PA, USA}
\icmlcorrespondingauthor{Shi Yu}{shi.yu@hotmail.com}
\icmlkeywords{Machine Learning }
\vskip 0.3in
]



\printAffiliationsAndNotice{\icmlEqualContribution} 

	\begin{abstract}
		We develop a chatbot using Deep Bidirectional Transformer models (BERT) \cite{devlin-etal-2019-bert} to handle client questions in financial investment customer service. The bot can recognize 381 intents, and decides when to say \textit{I don't know} and escalates irrelevant/uncertain questions to human operators. Our main novel contribution is the discussion about uncertainty measure for BERT, where three different approaches are systematically compared on real problems. We investigated two uncertainty metrics, \textit{information entropy} and \textit{variance of dropout sampling} in BERT, followed by mixed-integer programming to optimize decision thresholds. Another novel contribution is the usage of BERT as a language model in automatic spelling correction. Inputs with accidental spelling errors can significantly decrease intent classification performance. The proposed approach combines probabilities from masked language model and word edit distances to find the best corrections for misspelled words. The chatbot and the entire conversational AI system are developed using open-source tools, and deployed within our company's intranet. The proposed approach can be useful for industries seeking similar in-house solutions in their specific business domains. We share all our code and a sample chatbot built on a public dataset on Github.     
	\end{abstract}
	
	\section{Introduction}
	Since their first appearances decades ago \cite{ELIZA} \cite{Colby71}, Chatbots have always been marking the apex of Artificial Intelligence as forefront of all major AI revolutions, such as human-computer interaction, knowledge engineering, expert system, Natural Language Processing, Natural Language Understanding, Deep Learning, and many others. Open-domain chatbots, also known as \textit{Chitchat} bots, can mimic human conversations to the greatest extent in topics of almost any kind, thus are widely engaged for socialization, entertainment, emotional companionship, and marketing. Earlier generations of open-domain bots, such as Mitsuku\cite{Mitsuku} and ELIZA\cite{ELIZA}, relied heavily on hand-crafted rules and recursive symbolic evaluations to capture the key elements of human-like conversation. New advances in this field are mostly data-driven and end-to-end systems based on statistical models and neural conversational models \cite{Gao2018NeuralAT} aim to achieve human-like conversations through more scalable and adaptable learning process on free-form and large data sets \cite{Gao2018NeuralAT}, such as MILABOT\cite{DBLP:journals/corr/abs-1709-02349}, XiaoIce\cite{XiaoIce}, Replika\cite{Replika}, Zo\cite{Zo}, and Meena\cite{adiwardana2020humanlike}. 
	
	Unlike open-domain bots, closed-domain chatbots are designed to transform existing processes that rely on human agents. Their goals are to help users accomplish specific tasks, where typical examples range from order placement to customer support, therefore they are also known as \textit{task-oriented bots} \cite{Gao2018NeuralAT}. Many businesses are excited about the prospect of using closed-domain chatbots to interact directly with their customer base, which comes with many benefits such as cost reduction, zero downtime, or no prejudices. However, there will always be instances where a bot will need a human’s input for new scenarios. This could be a customer presenting a problem it has never expected for \cite{larson-etal-2019-evaluation}, attempting to respond to a naughty input, or even something as simple as incorrect spelling. Under these scenarios, expected responses from open-domain and closed-domain chatbots can be very different: a successful open-domain bot should be \textit{"knowledgeable, humourous and addictive"}, whereas a closed-domain chatbot ought to be \textit{"accurate, reliable and efficient"}. One main difference is the way of handling unknown questions. A chitchat bot would respond with an adversarial question such as \textit{Why do you ask this?}, and keep the conversation going and deviate back to the topics under its coverage \cite{2019bots}. A user may find the chatbot is out-smarting, but not very helpful in solving problems. In contrast, a task-oriented bot is scoped to a specific domain of intents, and should terminate out-of-scope conversations promptly and escalate them to human agents.              
	
	\begin{figure*}[h]
		\centering
		\includegraphics[width=\linewidth]{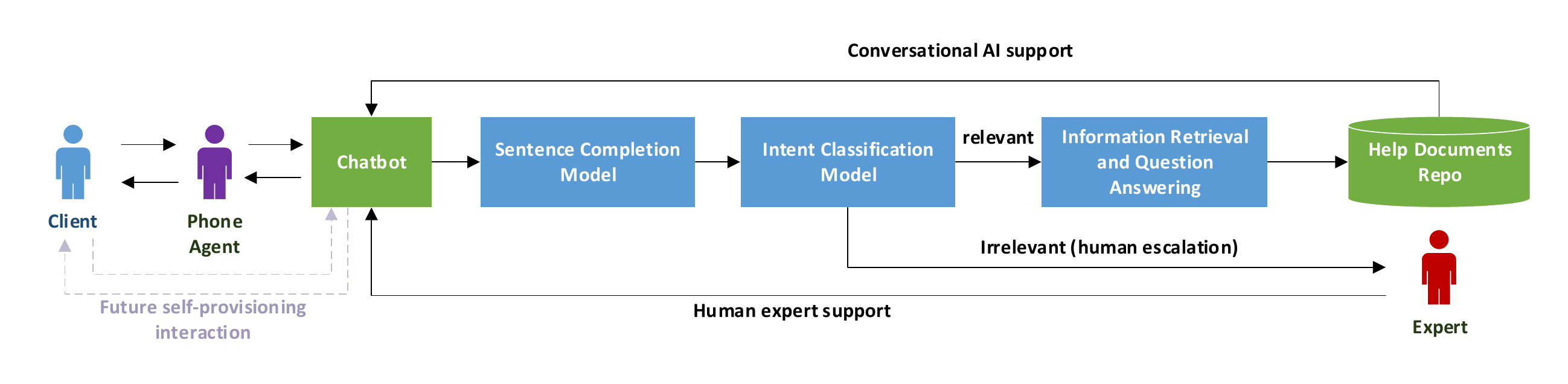}
		\caption{End-to-end conceptual diagram of AVA}
	\end{figure*}

	This paper presents AVA (\textbf{A} \textbf{V}anguard \textbf{A}ssistant), a task-oriented chatbot supporting phone call agents when they interact with clients on live calls. Traditionally, when phone agents need help, they put client calls on hold and consult experts in a support group. With a chatbot, our goal is to transform the consultation processes between phone agents and experts to an end-to-end conversational AI system. Our focus is to significantly reduce operating costs by reducing the call holding time and the need of experts, while transforming our client experience in a way that eventually promotes client self-provisioning in a controlled environment. Understanding intents correctly and escalating irrelevant intents promptly are keys to its success. Recently, NLP community has made many breakthroughs in context-dependent embeddings and bidirectional language models like ELMo, OpenAI, GPT, BERT, RoBERTa, DistilBERT, XLM, XLNet \cite{NIPS2015_5949, peters-etal-2017-semi, devlin-etal-2019-bert, peters-etal-2018-deep, DBLP:journals/corr/abs-1901-07291, DBLP:journals/corr/abs-1808-08949, howard-ruder-2018-universal, DBLP:journals/corr/abs-1906-08237, DBLP:journals/corr/abs-1907-11692, DBLP:journals/corr/abs-1903-12136}. In particular, the BERT model \cite{devlin-etal-2019-bert} has become a new NLP baseline including sentence classification, question answering, named-entity recognition and many others. To our knowledge there are few measures that address prediction uncertainties in these sophisticated deep learning structures, or explain how to achieve optimal decisions on observed uncertainty measures. The off-the-shelf softmax outputs of these models are predictive probabilities, and they are not a valid measure for the confidence in a network’s predictions \cite{Gal_2016, DBLP:journals/corr/abs-1902-02476, Pearce18,DBLP:journals/corr/abs-1901-02731}, which are important concerns in real-world applications \cite{larson-etal-2019-evaluation}.
	
	Our main contribution in this paper is applying advances in Bayesian Deep Learning to quantify uncertainties in BERT intent predictions. Formal methods like Stochastic Gradient (SG)-MCMC \cite{LiCVPR06, RaoSEFAIAS, WellingBayesian11, Park18, DBLP:journals/corr/abs-1902-02476, Seedat2019TowardsCA}, variational inference \cite{pmlr-v37-blundell15, Gal_2016, NIPS2011_4329, HernandezICML15} extensively discussed in literature may require modifying the network. Re-implementation of the entire BERT model for Bayesian inference is a non-trivial task, so here we took the Monte Carlo Dropout (MCD) approach \cite{Gal_2016} to approximate variational inference, whereby dropout is performed at training and test time, using multiple dropout masks. Our dropout experiments are compared with two other approaches (Entropy and Dummy-class), and the final implementation is determined among the trade-off between accuracy and efficiency. 
	
	We also investigate the usage of BERT as a language model to decipher spelling errors. Most vendor-based chatbot solutions embed an additional layer of service, where device-dependent error models and N-gram language models\cite{ngram} are utilized for spell checking and language interpretation.  At representation layer, Wordpiece model\cite{wordpiece} and Byte-Pair-Encoding(BPE) model\cite{BPE94,BPE} are common techniques to segment words into smaller units, thus similarities at sub-word level can be captured by NLP models and generalized on out-of-vocabulary(OOV) words. Our approach combines efforts of both sides: words corrected by the proposed language model are further tokenized by Wordpiece model to match pre-trained embeddings in BERT learning. 
	
    Despite all advances of chatbots, industries like finance and healthcare are concerned about cyber-security because of the large amount of sensitive information entered during chatbot sessions. Task-oriented bots often require access to critical internal systems and confidential data to finish specific tasks. Therefore, 100\% on-premise solutions that enable full customization, monitoring, and smooth integration are preferable than cloud solutions.  In this paper, the proposed chatbot is designed using RASA open-source version and deployed within our enterprise intranet.  Using RASA's conversational design, we hybridize RASA's chitchat module with the proposed task-oriented conversational systems developed on Python, Tensorflow and Pytorch.  We believe our approach can provide some useful guidance for industries contemplate adopting chatbot solutions in their business domains. 

	\begin{table*}
		\tiny
		\begin{tabular}{ c | c | c | l }
			\hline			
			T1 label & T2 label & T3 label & Questions \\ \hline 
			\multirow{3}{*}{Account Maintenance} & Call Authentication & Type 2  &  Am I allowed to give the client their Social security number?\\
			& Call Authentication & Type 5  &  Do the web security questions need to be reset by the client if their web access is blocked? \\
			& Web Reset  & Type 1 & How many security questions are required to be asked to reset a client’s web security questions? \\ \hline
			\multirow{2}{*}{Account Permission} & Call Authentication  & Type 2 &  How are the web security questions used to authenticate a client? \\
			& Agent Incapactiated &  Type 3 & Is it possible to set up Agent Certification for an Incapacitated Person on an Individual Roth 401k? \\ \hline
			TAX FAQ            & Miscellaneous &  What is & Do I need my social security number on the 1099MISC form? \\ \hline
			\multirow{2}{*}{Transfer of Asset} & Unlike registrations & Type 2 & Does the client need to provide special documentation if they want to transfer from one account to another account? \\
			& Brokerage Transfer & Type 3 & Is there a list of items that need to be included on a statement to transfer an account? \\ \hline
			\multirow{2}{*}{Banking}  &  Add Owner & Type 4 &  Once a bank has been declined how can we authorize it? \\
			&  Add/Change/Delete & Type 3 &  Does a limited agent have authorization to adjust bank info?  \\ \hline
			\multirow{3}{*}{Irrelevant} & - & - &  How can we get into an account with only one security question? \\
			& - & - &  Am I able to use my Roth IRA to set up a margin account? \\
			& - & - &  What is the best place to learn about Vanguard's investment philosophy? \\                        
			\hline  
		\end{tabular}
		\caption{Example questions used in AVA intent classification model training}
	\end{table*}
	
	\section{Background}
	Recent breakthroughs in NLP research are driven by two intertwined directions: Advances in distributed representations, sparked by the success of word embeddings \cite{Mikolov10,MikolovSCCD13}, character embeddings \cite{DBLP:journals/corr/KimJSR15,dos-santos-gatti-2014-deep, SantosICML14}, contextualized word embeddings \cite{peters-etal-2018-deep, radford, devlin-etal-2019-bert}, have successfully tackled the curse of dimensionality in modeling complex language models. Advances of neural network architecture, represented by CNN \cite{Collobert_CNN, DBLP:journals/corr/abs-1103-0398}, RNN\cite{Elman90findingstructure}, Attention Mechanism \cite{DBLP:journals/corr/BahdanauCB14}, and Transformer as seq2seq model with parallelized attentions \cite{Vaswani2017AttentionIA}, have defined the new state of the art deep learning models for NLP.
	
	Principled uncertainty estimation in regression \cite{Kuleshov_2018}, reinforcement learning \cite{Ghavamzadeh_2016} and classification \cite{Guo_2017} are active areas of research with a large volume of work. The theory of Bayesian neural networks \cite{Neal_1995, MacKay_1992} provides the tools and techniques to understand model uncertainty, but these techniques come with significant computational costs as they double the number of parameters to be trained. Gal and Ghahramani \cite{Gal_2016} showed that a neural network with dropout turned on at test time is equivalent to a deep Gaussian process and we can obtain model uncertainty estimates from such a network by multiple-sampling the predictions of the network at test time. Non-Bayesian approaches to estimating the uncertainty are also shown to produce reliable uncertainty estimates \cite{Lakshminarayanan_2017}; our focus in this paper is on Bayesian approaches. In classification tasks, the uncertainty obtained from multiple-sampling at test time is an estimate of the confidence in the predictions similar to the entropy of the predictions. In this paper, we compare the threshold for escalating a query to a human operator using model uncertainty obtained from dropout-based chatbot against setting the threshold using the entropy of the predictions. We choose dropout-based Bayesian approximation because it does not require changes to the model architecture, does not add parameters to train, and does not change the training process as compared to other Bayesian approaches. We minimize noise in the data by employing spelling correction models before classifying the input. Further, the labels for the user queries are human curated with minimal error. Hence, our focus is on quantifying epistemic uncertainty in AVA rather than aleatoric uncertainty \cite{kendall_2017}. We use mixed-integer optimization to find a threshold for human escalation of a user query based on the mean prediction and the uncertainty of the prediction. This optimization step, once again, does not require modifications to the network architecture and can be implemented separately from model training. In other contexts, it might be fruitful to have an integrated escalation option in the neural network \cite{Geifman_2019}, and we leave the trade-offs of integrated reject option and non-Bayesian approaches for future work. 
	
	Similar approaches in spelling correction, besides those mentioned in Section 1, are reported in Deep Text Corrector \cite{deeptextcorrector2017} that applies a seq2seq model to automatically correct small grammatical errors in conversational written English. Optimal decision threshold learning under uncertainty is studied in \cite{NIPS2016_6494} as Reinforcement learning and iterative Bayesian optimization formulations.  
	
	\section{System Overview and Data Sets}
	
	\subsection{Overview of the System}
	
	Figure 1 illustrates system overview of AVA. The proposed conversational AI will gradually replace the traditional human-human interactions between phone agents and internal experts, and eventually allows clients self-provisioning interaction directly to the AI system. Now, phone agents interact with AVA chatbot deployed on Microsoft Teams in our company intranet, and their questions are preprocessed by a Sentence Completion Model (introduced in Section 6) to correct misspellings. Then, inputs are classified by an intent classification model (Section 4 \& 5), where relevant questions are assigned predicted intent labels, and downstream information retrieval and questioning answering modules are triggered to extract answers from a document repository. Irrelevant questions are escalated to human experts following the decision thresholds optimized using methods introduced in section 5. This paper only discusses the Intent Classification model and the Sentence Completion model. 

	\subsection{Data for Intent Classification Model}

	Training data for AVA's intent classification model is collected, curated, and generated by a dedicated business team from interaction logs between phone agents and the expert team. The whole process takes about one year to finish. In total 22,630 questions are selected and classified to 381 intents, which compose the \textit{relevant questions set} for the intent classification model. Additionally, 17,395 questions are manually synthesized as \textit{irrelevant questions}, and none of them belongs to any of the aforementioned 381 intents. Each \textit{relevant question} is hierarchically assigned with three labels from Tier 1 to Tier 3. In this hierarchy, there are 5 unique Tier-1 labels, 107 Tier-2 labels, and 381 Tier-3 labels. Our intent classification model is designed to classify relevant input questions into 381 Tier 3 intents and then triggers downstream models to extract appropriate responses. The five Tier-1 labels and the numbers of intents include in each label are: \textit{Account Maintenance} (9074), \textit{Account Permissions} (2961), \textit{Transfer of Assets} (2838), \textit{Banking} (4788), \textit{Tax FAQ} (2969). At Tier-1, general business issues across intents are very different, but at Tier-3 level, questions are quite similar to each other, where differences are merely at the specific responses. Irrelevant questions, compared to relevant questions, have two main characteristics: 
	
	\begin{itemize}
		\item Some questions are relevant to business intents but unsuitable to be processed by conversational AI. For example, in Table 1, question \textit{"How can we get into an account with only one security question?"} is related to \textit{Call Authentication} in \textit{Account Permission}, but its response needs further human diagnosis to collect more information. These types of questions should be escalated to human experts.   
		\item Out of scope questions. For example, questions like \textit{"What is the best place to learn about Vanguard's investment philosophy?"} or \textit{"What is a hippopotamus?"} are totally outside the scope of our training data, but they may still occur in real world interactions.
	\end{itemize}

	\subsection{Textual Data for Pretrained Embeddings and Sentence Completion Model}
	Inspired by the progress in computer vision, transfer learning has been very successful in NLP community and has become a common practice. Initializing deep neural network with pre-trained embeddings, and fine-tune the models towards task-specific data is a proven method in multi-task NLP learning.  In our approach, besides applying off-the-shelf embeddings from Google BERT and XLNet, we also pre-train BERT embeddings using our company's proprietary text to capture special semantic meanings of words in the financial domain. Three types of textual datasets are used for embeddings training:     
	
	\begin{itemize}
		\item {Sharepoint text}: About 3.2G bytes of corpora scraped from our company's internal Sharepoint websites, including web pages, word documents, ppt slides, pdf documents, and notes from internal CRM systems. 
		\item {Emails}: About 8G bytes of customer service emails are extracted. 
		\item {Phone call transcriptions}: We apply AWS to transcribe 500K client service phone calls, and the transcription text is used for training.   
	\end{itemize}
	
	All embeddings are trained in case-insensitive settings. Attention and hidden layer dropout probabilities are set to 0.1, hidden size is 768, attention heads and hidden layers are set to 12, and vocabulary size is 32000 using SentencePiece tokenizer. On AWS P3.2xlarge instance each embeddings is trained for 1 million iterations, and takes about one week CPU time to finish. More details about parameter selection for pre-training are avaialble in the github code. The same pre-trained embeddings are used to initialize BERT model training in intent classification, and also used as language models in sentence completion.   
	
	\begin{figure}[h]%
		\begin{subfigure}
			\centering
			\includegraphics[width=\linewidth]{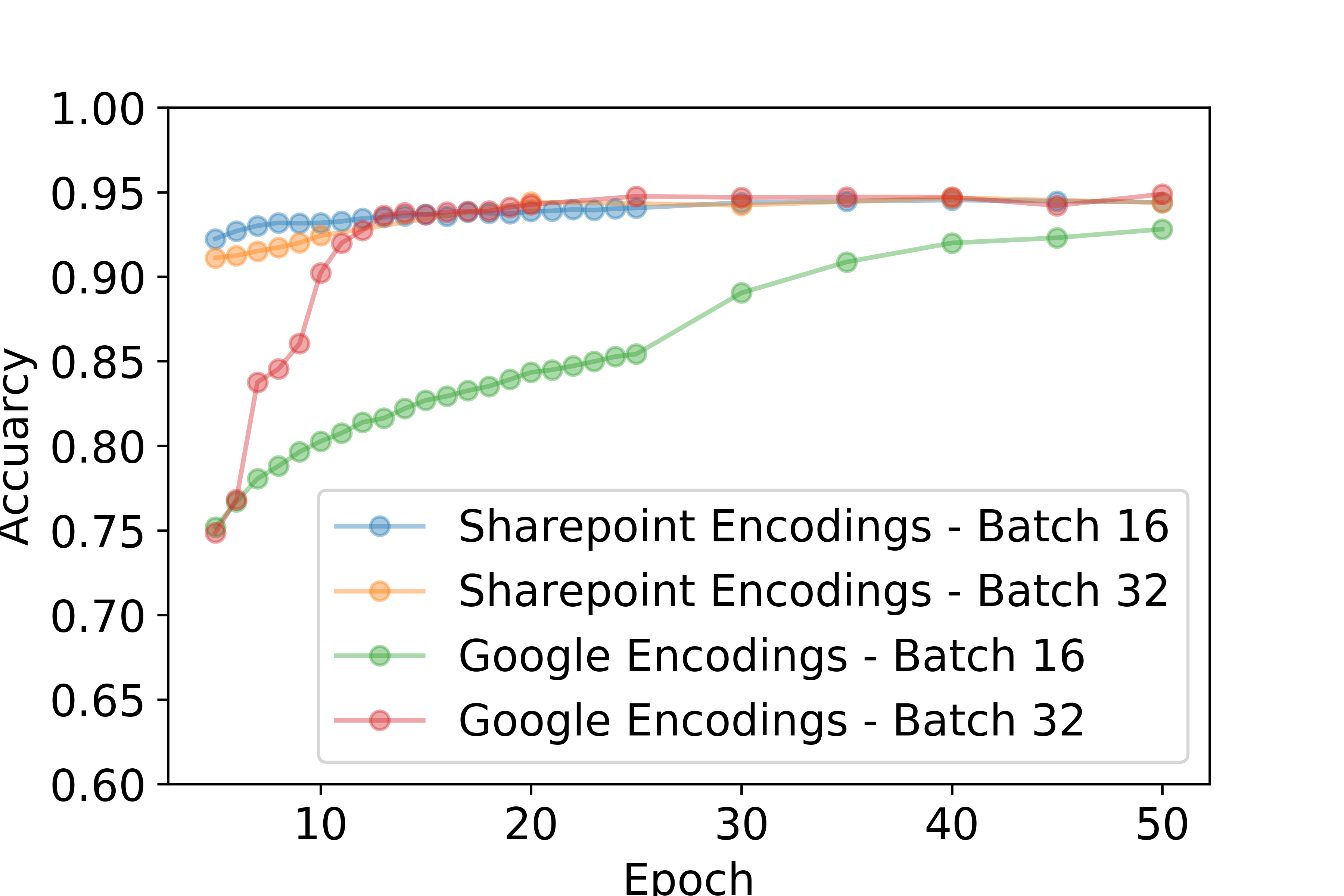}
		\end{subfigure}%
		\begin{subfigure}
			\centering
			\includegraphics[width=\linewidth]{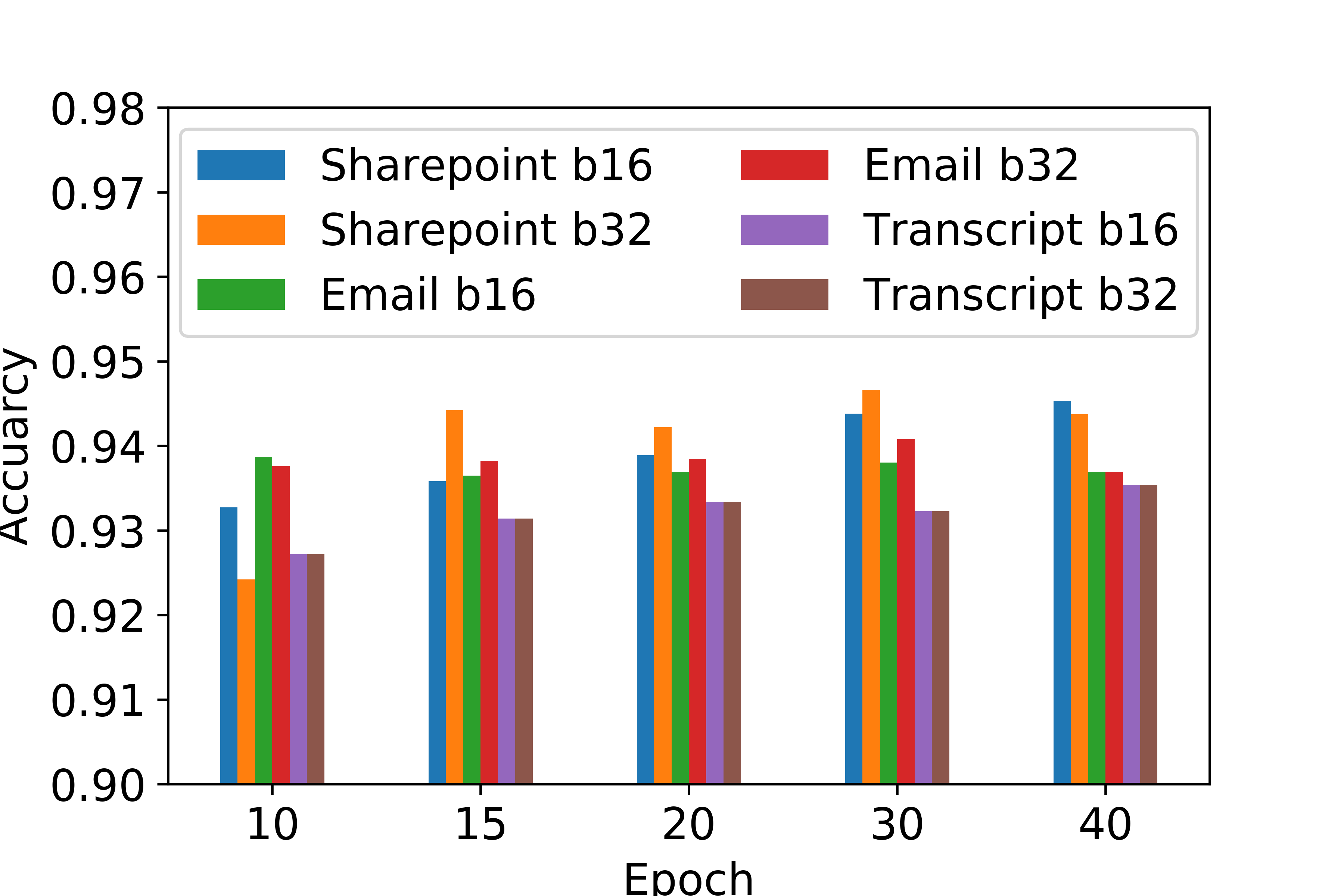}
		\end{subfigure}	
		\caption{Comparison of test set accuracy using different Embeddings and Batch Sizes}%
		\label{fig:example}%
	\end{figure}
	
    \begin{table}[h]
		\begin{tabular}{ c | c }
			\hline			
			Model & Performance \\ \hline 
			BERT small + Sharepoint Embeddings &  0.944  \\
			BERT small + Google Embeddings & 0.949 \\
			BERT large + Google Embeddings & 0.954 \\
			XLNet Large + Google Embeddings & 0.927 \\
			LSTM with Attention + Word2Vec  & 0.913 \\
			LSTM + Word2Vec  & 0.892 \\
			Logistic Regression + TFIDF & 0.820 \\
			Xgboost + TFIDF & 0.760 \\
			Naive Bayes + TFIDF & 0.661 \\  
			\hline  
		\end{tabular}
		\caption{Comparison of intent classification performance. BERT and XLNet models were all trained for 30 epochs using batch size 16.}
	\end{table}
	
	\section{Intent Classification Performance on Relevant Questions}
	
	Using only relevant questions, we compare various popular model architectures to find one with the best performance on 5-fold validation. Not surprisingly, BERT models generally produce much better performance than other models. Large BERT (24-layer, 1024-hidden, 16-heads) has a slight improvement over small BERT (12-layer, 768-hidden, 12-heads), but less preferred because of expensive computations. To our surprise, XLNet, a model reported outperforming BERT in mutli-task NLP, performs 2 percent lower on our data.   

	BERT models initialized by proprietary embeddings converge faster than those initialized by off-the-shelf embeddings (Figure 2.a). And embeddings trained on company's sharepoint text perform better than those built on Emails and phone-call transcriptions (Figure 2.b). Using larger batch size (32) enables models to converge faster, and leads to better performance.

	

	\section{Intent Classification Performance including Irrelevant Questions}
	We have shown how BERT model outperforming other models on real datasets that only contain relevant questions. The capability to handle 381 intents simultaneously at 94.5\% accuracy makes it an ideal intent classifier candidate in a chatbot. This section describes how we quantify uncertainties on BERT predictions and enable the bot to detect irrelevant questions. Three approaches are compared:
	
	\begin{itemize}
		\item {\textbf{Predictive-entropy}}: We measure uncertainty of predictions using Shannon entropy $H = -\sum_{k=1}^{K} p_{ik} \log{p_{ik}}$ where $p_{ik}$ is the prediction probability of $i$-th sample to $k$-th class. Here, $p_{ik}$ is softmax output of the BERT network \cite{Lakshminarayanan_2017}. A higher predictive entropy corresponds to a greater degree of uncertainty. Then, an optimally chosen cut-off threshold applied on entropies should be able to separate the majority of in-sample questions and irrelevant questions.
		\item {\textbf{Drop-out}}: We apply Monte Carlo (MC) dropout by doing 100 Monte Carlo samples. At each inference iteration, a certain percent of the set of units to drop out. This generates random predictions, which are interpreted as samples from a probabilistic
		distribution \cite{Gal_2016}. Since we do not employ regularization in our network, $\tau^{-1}$ in Eq. 7 in Gal and Ghahramani \cite{Gal_2016} is effectively zero and the predictive variance is equal to the sample variance from stochastic passes. We could then investigate the distributions and interpret model uncertainty as mean probabilities and variances.      
		\item {\textbf{Dummy-class}}: We simply treat escalation questions as a dummy class to distinguish them from original questions. Unlike entropy and dropout, this approach requires retraining of BERT models on the expanded data set including dummy class questions.   
	\end{itemize}
	
    \begin{figure*}[htp]
      \centering
      \subfigure[Predictive Entropy distributions on relevant (orange) and escalation (blue) questions]{\includegraphics[scale=0.2]{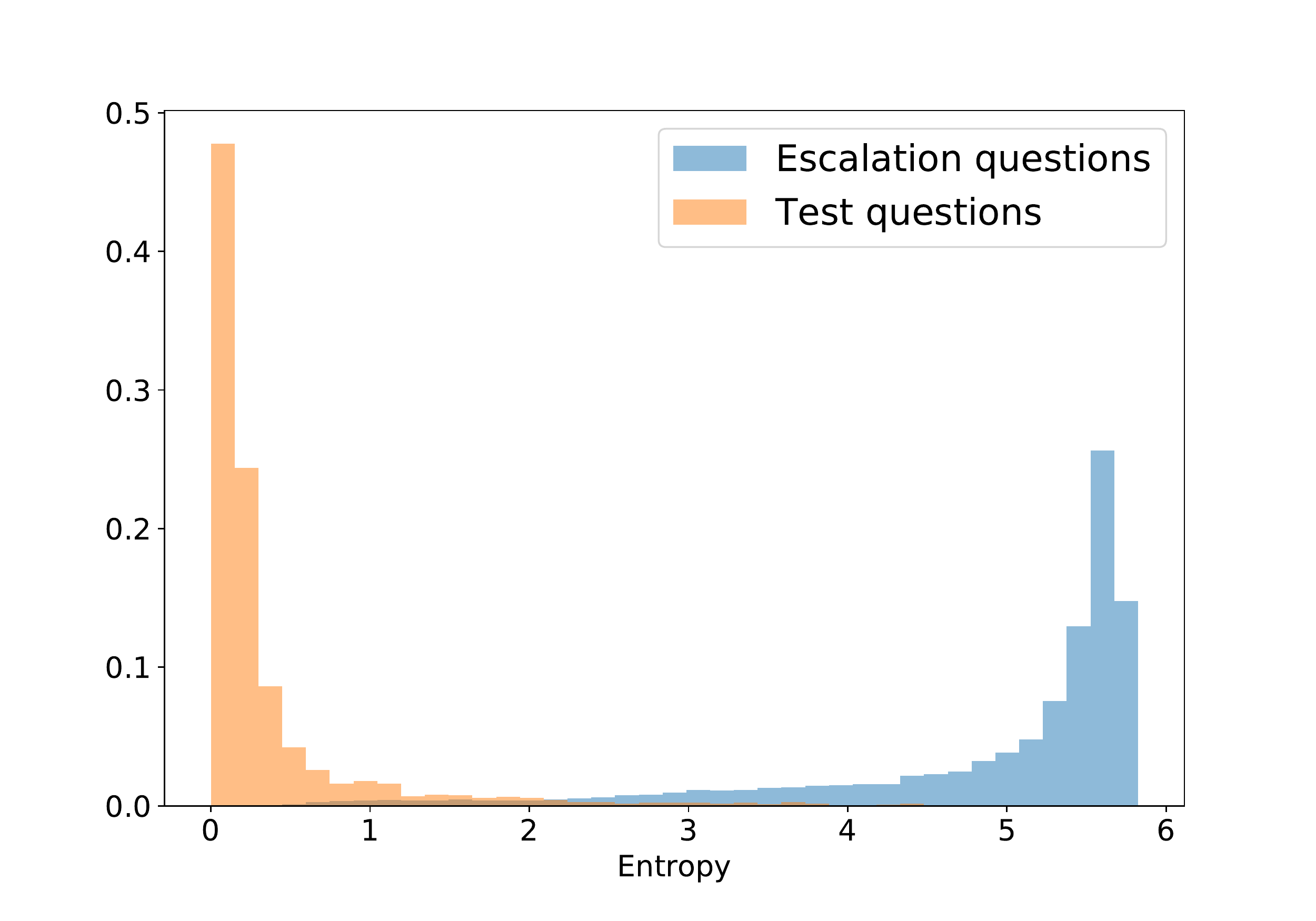}}\quad
      \subfigure[Test Accuracy when adding Escalation Questions in optimization procedure]{\includegraphics[scale=0.2]{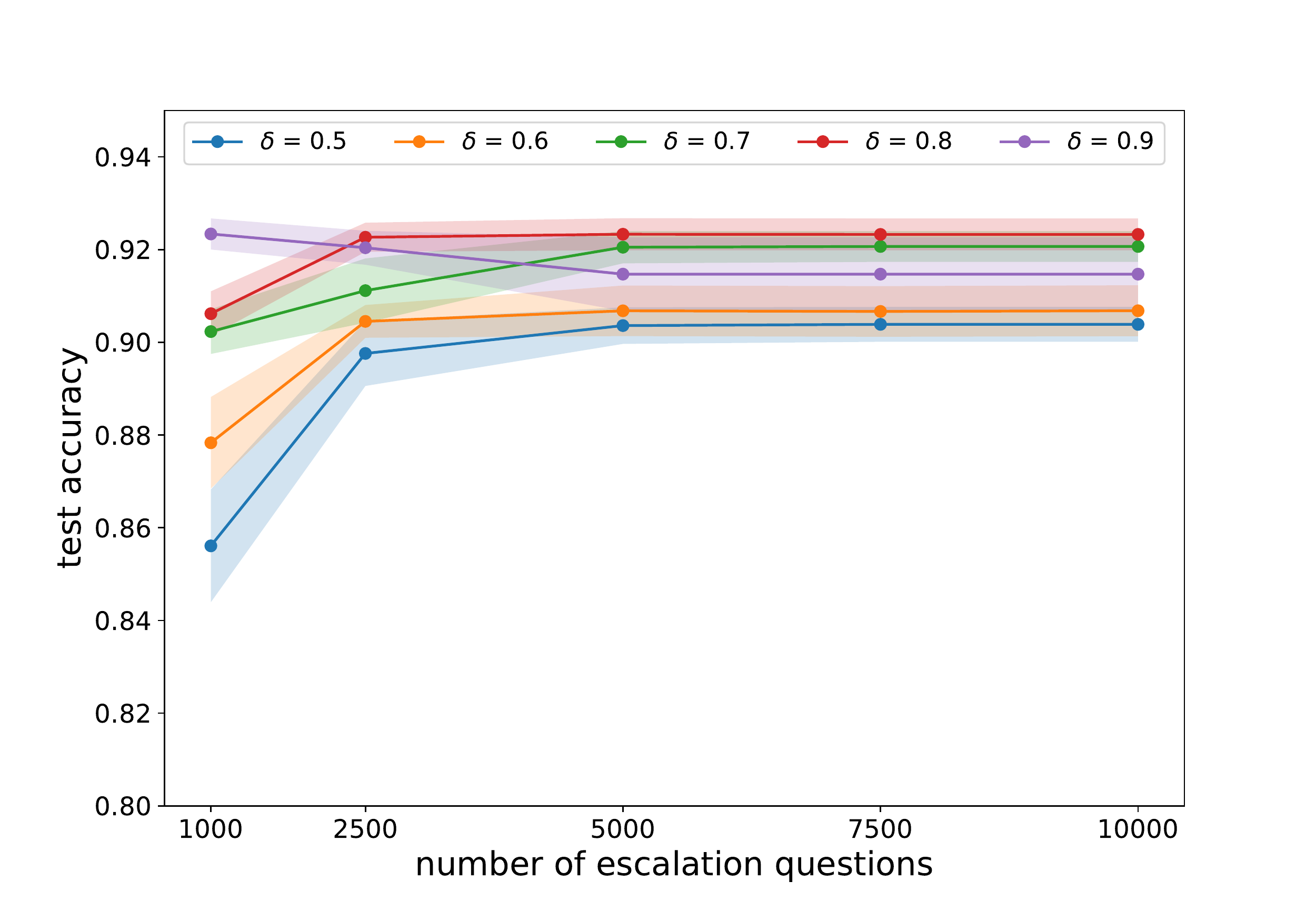}}\quad
      \subfigure[Optimal Entropy cut-off value when adding Escalation Questions in optimization procedure]{\includegraphics[scale=0.2]{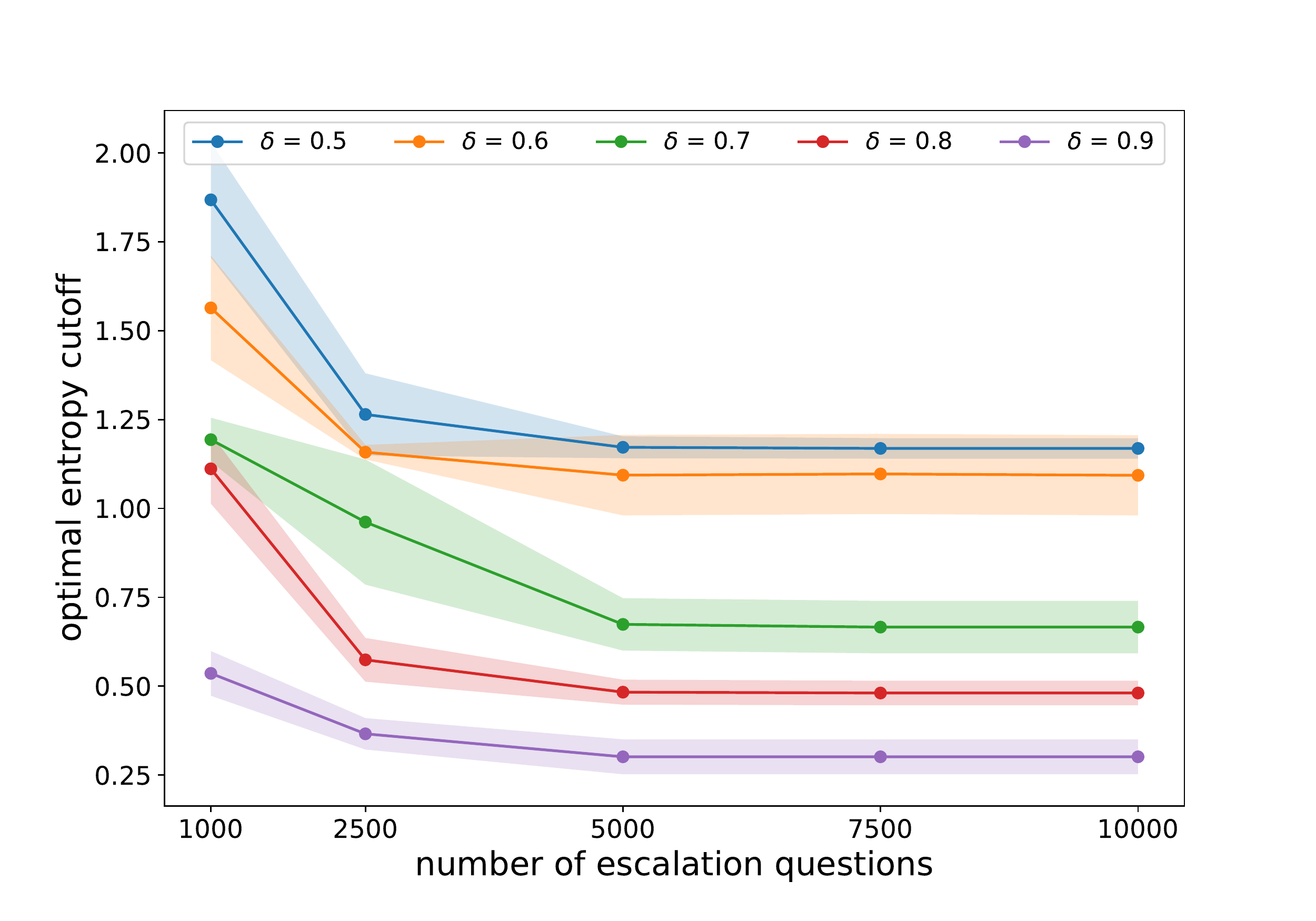}}
\caption{Optimizing entropy threshold to detect irrelevant questions. As shown in (a), in-sample test questions and irrelevant questions have very different distributions of predictive entropies. Subfigure (b) shows how test accuracies, evaluated using decision variables $b$ solved by (1) on BERT predictions on test data, change when different numbers of irrelevant questions involved in training. Subfigure (c) shows the impact of $\delta$ on the optimized thesholds when number of irrelevant questions increase optimization.}      
    \end{figure*}

	
	\subsection{Experimental Setup}
	All results mentioned in this section are obtained using BERT small + sharepoint embeddings (batch size 16). In Entropy and Dropout approaches, both relevant questions and irrelevant questions are split into five folds, where four folds (80\%) of relevant questions are used to train the BERT model. Then, among that 20\% held-out relevant questions we further split them into five folds, where 80\% of them (equals to 16\% of the entire relevant question set) are combined with four folds of irrelevant questions to learn the optimal decision variables. The learned decision variables are applied on BERT predictions of the remaining 20\% (906) of held-out relevant questions and held-out irrelevant questions (4000), to obtain the test performance. In dummy class approach, BERT model is trained using four folds of relevant questions plus four folds of irrelevant questions, and tested on the same amount of test questions as Entropy and Dropout approaches.
	
	\subsection{Optimizing Entropy Decision Threshold}

	To find the optimal threshold cutoff $b$, we consider the following Quadratic Mixed-Integer programming problem 
	\begin{align}
		\begin{array}{llll}   
			\displaystyle \min_{x, b} & \sum_{i,k}(x_{ik}-l_{ik})^2 \\
			\textrm{s.t.} 
			& x_{ik}= 0 & \textrm{if } E_{i} \geq b, \textrm{ for } k \textrm{ in } 1,...,K \quad \\
			& x_{ik}= 1 & \textrm{if } E_{i} \geq b, \textrm{ for } k=K+1 \quad \\
			& x_{ik} \in \{0,1\}  \\
			& \sum_{k=1}^{K+1} x_{ik} = 1 &  \forall i \textrm{ in } 1,...,N \\
			& b \geq 0
		\end{array}
	\end{align}
	\noindent to minimize the quadratic loss between the predictive assignments $x_{ik}$ and true labels $l_{ik}$. In (1), $i$ is sample index, and $k$  is class (intent) indices. $x_{ik}$ is $N \times (K+1)$ binary matrix, and $l_{ik}$ is also $N \times (K+1)$, where the first $K$ columns are binary values and the last column is a uniform vector $\delta$, which represents the cost of escalating questions. Normally $\delta$ is a constant value smaller than 1, which encourages the bot to escalate questions rather than making mistaken predictions. The first and second constraints of (1) force an escalation label when entropy $E_{i} \geq b$. The third and fourth constraints restrict $x_{ik}$ as binary variables and ensure the sum for each sample is 1. Experimental results (Figure 3) indicate that (1) needs more than 5000 escalation questions to learn a stabilized $b$. The value of escalation cost $\delta$ has a significant impact on the optimal $b$ value,  and in our implementation is set to 0.5. 

\begin{figure*}[h]
  \centering
  \subfigure[Intent accuracy at different drop out ratios]{\includegraphics[scale=0.26]{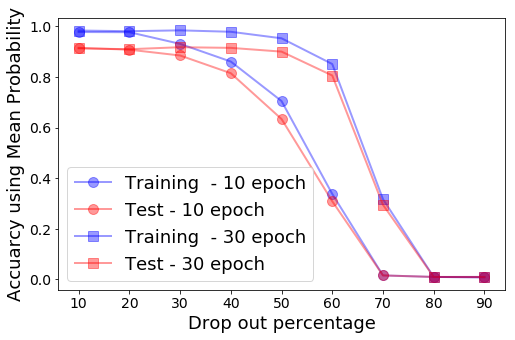}}\quad
  \subfigure[Uncertainties on questions after training 10 epochs]{\includegraphics[scale=0.26]{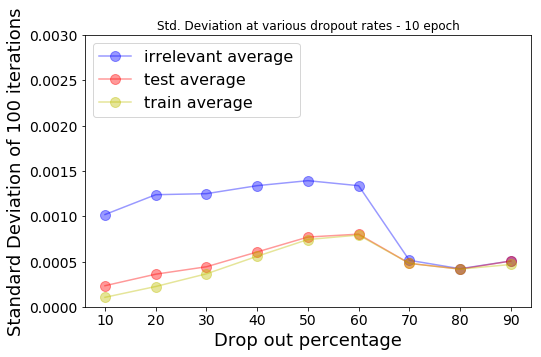}}\quad
  \subfigure[Uncertainties on questions after training 30 epochs]{\includegraphics[scale=0.26]{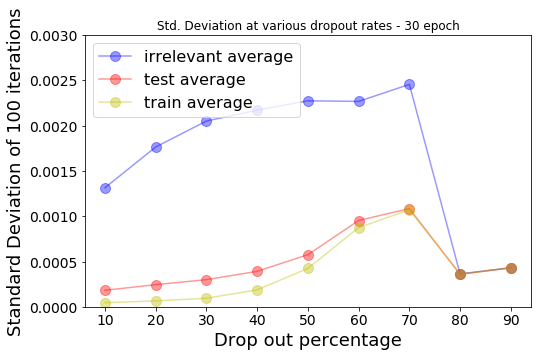}}
  \caption{Classification accuracy and uncertainties obtained from Monte Carlo Dropout}
\end{figure*}	    
	
	
	\begin{table*}
		\tiny
		\begin{tabular}{ | l || c | c | c | c || c | c | c | c || c | c | c | c | }
			\hline 
			& \multicolumn{4}{|c||}{Entropy} & \multicolumn{4}{c||}{Dropout} & \multicolumn{4}{c|}{Dummy Class} \\
			\hline
			number of irrelevant  & & & & & & & & & & & &\\
			questions in training    & 1000 & 5000 & 8000 & 10000 & 100    & 1000  & 2000  & 3000  &  1000 &   5000 & 8000 & 10000 \\ \hline
			optimal entropy cutoff $b$  & 2.36 & 1.13 & 0.85 & 0.55  &    -   & -  &-   &-      & -     & -      &  -    &  -    \\   \hline 
			optimal mean prob cutoff $c$&  -  & - & - & -  & 0.8172 &0.6654 &0.7921 & 0.0459& -     &  -     &  -&  -     \\  \hline 
			optimal std. cutoff  $d$  &  - &  -  &- & - & 0.1533 &0.0250 &0.0261 & 0.0132&- &- &- &- \\  \hline
			mean accuracy in $381$ classes & $91.9\%$ & $88.3\%$ & $85.6\%$ & $81.7\%$ & $88.41\%$ & $80.13\%$  & $80.24\%$ & $74.72\%$ & $94.2\%$ & \textbf{93.7\%} & $87.7\%$ & $82\%$ \\ \hline
			accuracy of the dummy class & $79.25\%$ & $91.2\%$ & $93.25\%$ & $95.2\%$ & $86.69\%$ & $91.83\%$ & $91.95\%$ & $92.57\%$ & $73.6\%$ & \textbf{94.5\%} & $99.4\%$ & $99.6\%$ \\
			\hline
			precision (binary classification) & 51.4\% & 70.2\% & 74.7\% & 79.8\% & 90.7\% & 68.8\% & 68.9\% & 63.7\% & 81\% & \textbf{95.3\%} & 99.5\% & 99.6\%\\ \hline
			recall (binary classification) & 96.7\% & 91.3\% & 88.1\%& 83.5\% & 93.9\% & 82.7\% & 83.2\% & 84.7\% & 99.7\% & \textbf{98.7\%} & 92.6\% & 86\% \\ \hline
			F1 score (binary classification) & 0.671 & 0.794 & 0.808 & 0.816 & 0.738 & 0.751 & 0.754 & 0.727 &  0.894 & \textbf{0.967} & 0.959 & 0.923 \\ \hline
		\end{tabular}
		\caption{Performance cross comparison of three approaches evaluated on test data of same size (906 relevant questions plus 4000 irrelevant questions).  Precision/Recall/F1 scores were calculated assuming relevant questions are true positives. In entropy and dropout optimization processes, $\delta$ is set to 0.5. Other delta values for dropout approach are listed in appendix. }
	\end{table*}

	\subsection{Monte Carlo Drop-out}
	In BERT model, dropout ratios can be customized at encoding, decoding, attention, and output layer. A combinatorial search for optimal dropout ratios is computationally challenging. Results reported in the paper are obtained through simplifications with the same dropout ratio assigned and varied on all layers. Our MC dropout experiments are conducted as follows:
	
	\begin{enumerate}
		\item [1] Change dropout ratios in encoding/decoding/attention/output layer of BERT
		\item [2] Train BERT model on 80\% of relevant questions for 10 or 30 epochs 
		\item [3] Export and serve the trained model by Tensorflow serving
		\item [4] Repeat inference 100 times on questions, then average the results per each question to obtain mean probabilities and standard deviations, then average the deviations for a set of questions.     
	\end{enumerate} 
	
	According to the experimental results illustrated in Figure 4, we make three conclusions: (1) Epistemic uncertainty estimated by MCD reflects question relevance: when inputs are similar to the training data there will be low uncertainty, whilst data is different from the original training data should have higher epistemic uncertainty. (2) Converged models (more training epochs) should have similar uncertainty and accuracy no matter what drop ratio is used. (3) The number of epochs and dropout ratios are important hyper-parameters that have substantial impacts on uncertainty measure and predictive accuracy and should be cross-validated in real applications.
	
	We use mean probabilities and standard deviations obtained from models where dropout ratios are set to 10\% after 30 epochs of training to learn optimal decision thresholds. Our goal is to optimize lowerbound $c$ and upperbound $d$, and designate a question as relevant only when the mean predictive probability $P_{ik}$ is larger than $c$ and standard deviation $V_{ik}$ is lower than $d$. Optimizing $c$ and $d$, on a 381-class problem, is much more computationally challenging than learning entropy threshold because the number of constraints is proportional to class number. As shown in (2), we introduce two variables $\alpha$ and $\beta$ to indicate the status of mean probability and deviation conditions, and the final assignment variables $x$ is the logical AND of $\alpha$ and $\beta$.  Solving (2) with more than 10k samples is very slow (shown in Appendix), so we use 1500 original relevant questions, and increase the number of irrelevant questions from 100 to 3000. For performance testing, the optimized $c$ and $d$ are applied as decision variables on samples of BERT predictions on test data. Performance from dropout are presented in Table 3 and Appendix. Our results showed decision threshold optimized from (2) involving 2000 irrelevant questions gave the best F1 score (0.754), and we validated it using grid search and confirmed its optimality (shown in appendix). 
	\begin{align}
		\begin{array}{lllll}
			\displaystyle \min_{x, c,d} & \sum_{i,k}(x_{ik}-l_{ik})^2\\
			\textrm{s.t.} 
			& \alpha_{ik}=\begin{cases}
				0 \quad & \textrm{if  } P_{ik} \leq c, \textrm{ for } k \textrm{ in } 1,...,K \quad\\
				1 \quad & \textrm{if  } \textrm{otherwise}   \\
			\end{cases}\\
			& \beta_{ik}=\begin{cases}
				0 \quad & \textrm{if  } V_{ik} \geq d, \textrm{ for } k \textrm{ in } 1,...,K \quad \\
				1 \quad & \textrm{if  } \textrm{otherwise}   \\
			\end{cases}\\
			& x_{ik}= 0 \quad   \textrm{if  } \alpha_{ik}=0 \textrm{  OR  } \beta_{ik}=0 \\
			& x_{ik}= 1 \quad   \textrm{if  } \alpha_{ik}=1 \textrm{  AND  } \beta_{ik}=1 \\
			& \sum_{k}^{K+1}x_{ik} = 1 \quad  \forall i \textrm{ in } 1,...,N \\
			& 1 \geq c \geq 0 \\
			& 1 \geq d \geq 0
		\end{array}
	\end{align}
	
			
	
	\subsection{Dummy-class Classification}
	Our third approach is to train a binary classifier using both relevant questions and irrelevant questions in BERT. We use a dummy class to represent those 17,395 irrelevant questions, and split the entire data sets, including relevant and irrelevant, into five folds for training and test. 

	
	Performance of dummy class approach is compared with Entropy and Dropout approaches (Table 3). Deciding an optimal number of irrelevant questions involved in threshold learning is non-trivial, especially for Entropy and Dummy class approaches. Dropout doesn't need as many irrelevant questions as entropy does to learn optimal threshold, mainly because the number of constraints in (2) is proportional to the class number (381), so the number of constraints are large enough to learn a suitable threshold on small samples (To support this conclusion, we present extensive studies in Appendix on a 5-class classifier using Tier 1 intents). Dummy class approach obtains the best performance, but its success assumes the learned decision boundary can be generalized well to any new irrelevant questions, which is often not valid in real applications. In contrast, Entropy and Dropout approaches only need to treat a binary problem in the optimization and leave the intent classification model intact. The optimization problem for entropy approach can be solved much more efficiently, and is selected as the solution for our final implementation.  
	
    It is certainly possible to combine Dropout and Entropy approach, for example, to optimize thresholds on entropy calculated from the average mean of MCD dropout predictions. Furthermore, it is possible that the problem defined in (2) can be simplified by proper reformulation, and can be solved more efficiently, which will be explored in our future works.

	\section{Sentence Completion using Language Model}
	
	\subsection{Algorithm}
	 We assume misspelled words are \textbf{all} OOV words, and we can transform them as [MASK] tokens and use bidirectional language models to predict them. Predicting masked word within sentences is an inherent objective of a pre-trained bidirectional model, and we utilize the Masked Language Model API in the Transformer package \cite{Huggingface2017} to generate the ranked list of candidate words for each [MASK] position. The sentence completion algorithm is illustrated in Algorithm 1. 
	
	
	\begin{figure}[h]
	\centering
	\includegraphics[scale=0.5]{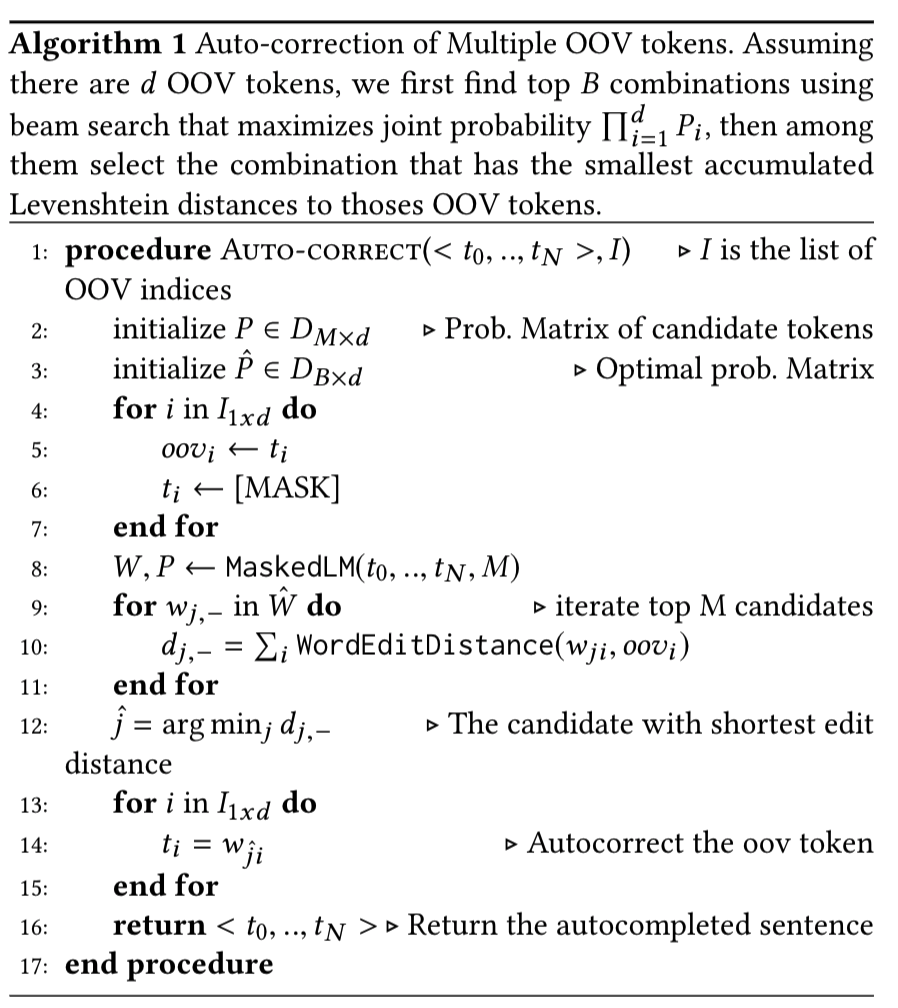}
	\end{figure}

			
			
			
			
	
\subsection{Experimental Setup}
	
¸\begin{figure*}[h]
  \centering
  \subfigure[Accuracy - Single OOV]{\includegraphics[scale=0.25]{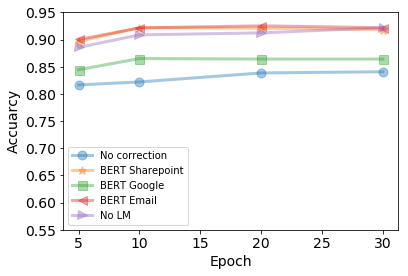}}\quad
  \subfigure[Accuracy - Two OOVs]{\includegraphics[scale=0.25]{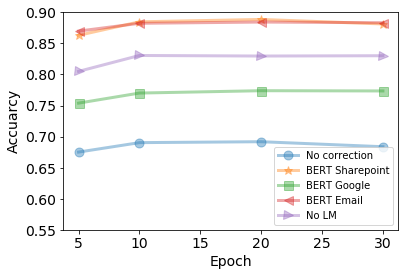}}\quad
  \subfigure[Accuracy - Three OOVs]{\includegraphics[scale=0.25]{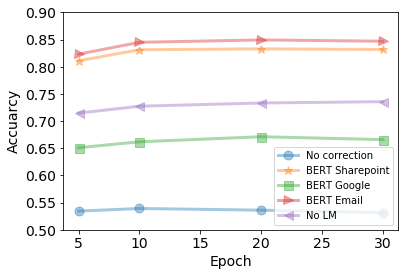}}\quad
  \subfigure[Accuracy per beam size]{\includegraphics[scale=0.25]{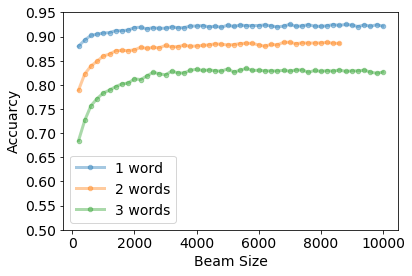}}  
  \caption{As expected, misspelled words can significantly decrease intent classification performance. The same BERT model that achieved 94\% on clean data, dropped to 83.5\% when a single OOV occured in each question. It further dropped to 68\% and 52\%, respectively, when two and three OOVs occured. In all experiments, LM models proved being useful to help correcting words and reduce performance drop, while domain specific embeddings trained on Vanguard Sharepoint and Email text outperform off-the-shelf Google embeddings. The beam size $B$ ($M$) was benchmarked as results shown in subfigure (d), and was set to 4000 to generate results in subfigure (a) to (c).}
\end{figure*}

	For each question, we randomly permutate two characters in the longest word, the next longest word, and so on. In this way, we generate one to three synthetic misspellings in each question. We investigate intent classification accuracy changes on these questions, and how our sentence completion model can prevent performance changes. All models are trained using relevant data (80\%) without misspellings and validated on synthetic misspelled test data. Five settings are compared: (1) \textbf{No correction}: classification performance without applying any auto-correction; (2) \textbf{No LM}: Auto-corrections made only by word edit distance without using Masked Language model; (3) \textbf{BERT Sharepoint}: Auto-corrections made by Masked LM using pre-trained sharepoint embeddings together with word edit distance;  (4) \textbf{BERT Email}: Auto-corrections using pretrained email embeddings together with word edit distance; (5) \textbf{BERT Google}: Auto-corrections using pretrained Google Small uncased embedding data together with word edit distance.
	
	We also need to decide what is an OOV, or, what should be included in our vocabulary. After experiments, we set our vocabulary as words from four categories: (1) All words in the pre-trained embeddings; (2) All words that appear in training questions; (3) Words that are all capitalized because they are likely to be proper nouns, fund tickers or service products; (4) All words start with numbers because they can be tax forms or specific products (e.g., 1099b, 401k, etc.). The purposes of including (3) and (4) is to avoid auto-correction on those keywords that may represent significant intents. Any word falls outside these four groups is considered as an OOV. During our implementation, we keep monitoring OOV rate, defined as the ratio of OOV occurrences to total word counts in recent 24 hours. When it is higher than 1\%, we apply manual intervention to check chatbot log data.     

	We also need to determine two additional parameters $M$, the number of candidate tokens prioritized by masked language model and $B$, the beam size in our sentence completion model. In our approach, we set $M$ and $B$ to the same value, and it is benchmarked from 1 to 10k by test sample accuracy. Notice that when $M$ and $B$ are large, and when there are more than two OOVs, Beam Search becomes very inefficient in Algorithm 1. To simplify this, instead of finding the optimal combinations of candidate tokens that maximize the joint probability $\arg\max \prod_{i=1}^{d} p_{i}$, we assume they are independent and apply a simplified Algorithm (shown in Appendix) on single OOV separately. An improved version of sentence completion algorithm to maximize joint probability will be our future research. We haven't consider situations when misspellings are not OOV in our paper. To detect improper words in a sentence may need evaluation of metrics such as Perplexity or Sensibleness and Specificity Average (SSA)\cite{adiwardana2020humanlike}, and will be our future goals.      
	
	\subsection{Results}
	
	
	
	According to the experimental results illustrated in Figure 5, pre-trained embeddings are useful to increase the robustness of intent prediction on noisy inputs. Domain-specific embeddings contain much richer context-dependent semantics that helps OOVs get properly corrected, and leads to better task-oriented intent classification performance. Benchmark shows B$\geq$4000 leads to the best performance for our problem. Based on this, we apply sharepoint embeddings as the language model in our sentence completion module. 
	
	\section{Implementation}
	The chatbot has been implemented fully inside our company network using open source tools including RASA\cite{RASA}, Tensorflow, Pytorch in Python enviornment. All backend models (Sentence Completion model, Intent Classification model and others) are deployed as RESTFUL APIs in AWS Sagemaker. The front-end of chatbot is launched on Microsoft Teams, powered by Microsoft Botframework and Microsoft Azure directory, and connected to backend APIs in AWS environment. All our BERT model trainings, including embeddings pretraining, are based on BERT Tensorflow running on AWS P3.2xlarge instance. The optimization procedure uses Gurobi 8.1 running on AWS C5.18xlarge instance. BERT language model API in sentence completion model is developed using Transformer 2.1.1 package on PyTorch 1.2 and Tensorflow 2.0. 
	
	During our implementation, we further explore how the intent classification model API can be served in real applications under budget. We gradually reduce the numbers of attention layer and hidden layer in the original BERT Small model (12 hidden layers, 12 attention heads) and create several smaller models. By reducing the number of hidden layers and attention layers in half, we see a remarkable 100\% increase in performance (double the throughput, half the latency) with the cost of only 1.6\% drop in intent classification performance.  
	
	\begin{table}[htb]
	\footnotesize
		\begin{tabular}{ c | c | c | c }
			\hline			
			Model              & Performance  & Throughput & Avg. Latency \\ \hline 
			
			12A-12H  &  0.944       & 8.9/s      &  1117 ms     \\
            6A-12H             &  0.941       & 9.0/s      &  1108 ms     \\
            12A-9H             &  0.934       & 11.8/s     &  843 ms      \\
            3A-9H              &  0.933       & 12.0/s     &  831 ms      \\
            3A-12H             &  0.930       & 9.1/s      &  1097 ms      \\
            6A-6H              &  0.928       & 18.1/s     &  552 ms      \\
			\hline  
		\end{tabular}
		\caption{Benchmark of intent classification API performance across different models in real application. Each model is tested using 10 threads, simulating 10 concurrent users, for a duration of 10 minutes. In this test, models are not served as Monte Carlo sampling, so the inference is done only once.  All models are hosted on identical AWS m5.4xlarge CPU instances. As seen, the simplest model (6A-6H, 6 attention layers and 6 hidden layers) can have double throughput rate and half latency than the original BERT small model, and the accuracy performance only drops 1.6\%. The performance is evaluated using JMeter at client side, and APIs are served using Domino Lab 3.6.17 Model API. Throughput indicates how many API responses being made per second.  Latency is measured as time elapse between request sent till response received at client side.}
	\end{table}	
	
	\section{Conclusions}
	 Our results demonstrate that optimized uncertainty thresholds applied on BERT model predictions are promising to escalate irrelevant questions in task-oriented chatbot implementation, meanwhile the state-of-the-art deep learning architecture provides high accuracy on classifying into a large number of intents. Another feature we contribute is the application of BERT embeddings as language model to automatically correct small spelling errors in noisy inputs, and we show its effectiveness in reducing intent classification errors. The entire end-to-end conversational AI system, including two machine learning models presented in this paper, is developed using open source tools and deployed as in-house solution. We believe those discussions provide useful guidance to companies who are motivated to reduce dependency on vendors by leveraging state-of-the-art open source AI solutions in their business.
	
	We will continue our explorations in this direction, with particular focuses on the following issues: (1) Current fine-tuning and decision threshold learning are two separate parts, and we will explore the possibility to combine them as a new cost function in BERT model optimization. (2) Dropout methodology applied in our paper belongs to approximated inference methods, which is a crude approximation to the exact posterior learning in parameter space. We are interested in a Bayesian version of BERT, which requires a new architecture based on variational inference using tools like TFP Tensorflow Probability. (3) Maintaining chatbot production system would need a complex pipeline to continuously transfer and integrate features from deployed model to new versions for new business needs, which is an uncharted territory for all of us. (4) Hybridizing "chitchat" bots, using state-of-the-art progresses in deep neural models, with task-oriented machine learning models is important for our preparation of client self-provisioning service.          
	
	\section{Acknowledgement}
	We thank our colleagues in Vanguard CAI (ML-DS team and IT team) for their seamless collaboration and support. We thank colleagues in Vanguard Retail Group (IT/Digital, Customer Care) for their pioneering effort collecting and curating all the data used in our approach. We thank Robert Fieldhouse, Sean Carpenter, Ken Reeser and Brain Heckman for the fruitful discussions and experiments.

\bibliography{chatbot}
\bibliographystyle{icml2019}

	\clearpage
	\appendix
	\section{Appendix}
	All extended materials and source code related to this paper are avaliable on \url{https://github.com/cyberyu/ava}    Our repo is composed of two parts: (1) Extended materials related to the main paper, and (2) Source code scripts. To protect proprietary intellectual property, we cannot share the question dataset and proprietary embeddings.  We use an alternative data set from Larson \textit{et al.}, \textit{``An Evaluation Dataset for Intent Classification and Out-of-Scope Prediction''}, EMNLP-IJCNLP 2019, to demonstrate the usage of code. 
	
	\subsection{Additional Results for the Main Paper}
	Some extended experimental results about MC dropout and optimization are presented on github. 
	
	\subsubsection{Histogram of Uncertainties by Dropout Ratios}
	
    We compare histograms of standard deviations observed from random samples of predictions.  The left side contains histograms generated by 381-class intent models trained for 10 epochs, with dropout ratio varied from 10 percent to 90 percent.  The right side shows histograms generated by 381 class models trained for 30 epochs. 

	\subsubsection{Uncertainty comparison between 381-class vs 5-class}
	To understand how uncertainties change vs. the number of classes in BERT, we train another intent classifier using only Tier 1 labels. We compare  uncertainty and accuracy changes at different dropout rates between the original 381-class problem and the new 5-class problem. 
	
	\subsubsection{Grid Search For Optimal Threshold on Dropout}
    Instead of using optimization, we use a grid search to find optimal combinations of average probability threshold and standard deviation threshold. The search space is set as a 100 x 100 grid on space [0,0] to [1,1], where thresholds vary by step of 0.01 from 0 to 1. Applying thresholds to outputs of BERT predictions give us classifications of relevance vs. irrelevance questions, and using the same combination of test and irrelevant questions we visualize the F1 score in contour map shown on github repo.   
	
	\subsubsection{Optimal Threshold Learning on Dropout 381 classes vs 5 classes} Using the same optimization process mentioned in equation (2) of the main paper, we compare the optimal results (also CPU timing) learned from 381 classes vs. 5 classes. 
	
	\subsubsection{Simple algorithm for sentence completion model} When multiple OOVs occur in a sentence, in order to avoid the computational burden using large beamsize to find the optimal joint probabilities, we assume all candidate words for OOVs are independent, and apply Algorithm 2 one by one to correct the OOVs.  
	
	\begin{figure}[h]
	\centering
	\includegraphics[scale=0.5]{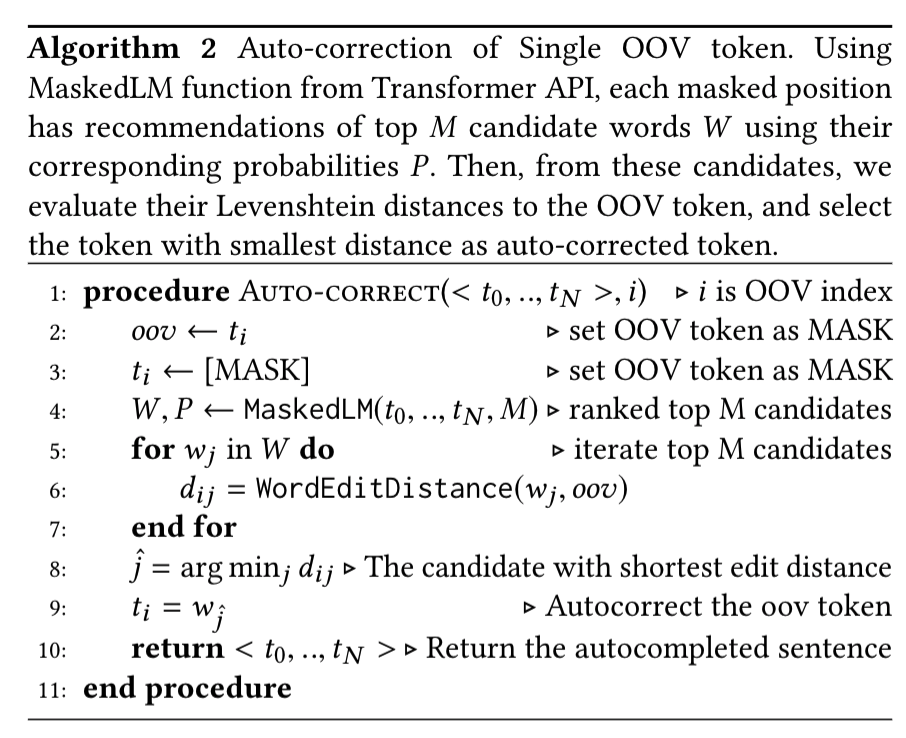}
	\end{figure}
	

	\subsection{Intent Classification Source Code}
	\subsubsection{BERT embeddings Model Pretraining}
	The jupyter notebook for pretraining embeddings is at \url{https://github.com/cyberyu/ava/blob/master/scripts/notebooks/BERT_PRETRAIN_Ava.ipynb}. Our script is adapted from Denis Antyukhov's blog ``Pre-training BERT from scratch with cloud TPU''. We set the \textrm{VOC\_SIZE} to 32000, and use SentencePiece tokenizer as approximation of Google's WordPiece. The learning rate is set to 2e-5, training batch size is 16, training setps set to 1 million, \textrm{MAX\_SEQ\_LENGTH} set to 128, and \textrm{MASKED\_LM\_PROB} is set to 0.15.
	
	To ensure the embeddings is training at the right architecture, please make sure the \textrm{bert\_config.json} file referred in the script has the right numbers of hidden and attention layers. 

	\subsubsection{BERT model training and exporting} The jupyter notebook for BERT intent classification model training, validation, prediciton and exporting is at \url{https://github.com/cyberyu/ava/blob/master/scripts/notebooks/BERT_run_classifier_Ava.ipynb}. The main script  \textrm{run\_classifier\_inmem.py} is tweaked from the default BERT script \textrm{run\_classifier.py}, where a new function \textrm{serving\_input\_fn():} is added.  To export that model in the same command once training is finished, the '--do\_export=true' need be set True, and the trained model will be exported to directory specified in '--export\_dir' FLAG.
	
	\subsubsection{Model Serving API Script} We create a jupyter notebook to demonstrate how exported model can be served as in-memory classifier for intent classification, located at \url{https://github.com/cyberyu/ava/scripts/notebooks/inmemory_intent.ipynb}. The script will load the entire BERT graph in memory from exported directory, keep them in memory and provide inference results on new questions. Please notice that in ``getSess()'' function, users need to specify the correct exported directory, and the correct embeddings vocabulary path. 
	
	\subsubsection{Model inference with Dropout Sampling} We provide a script that performs Monte Carlo dropout inference using in-memory classifier. The script assumes three groups of questions are saved in three separate files: training.csv, test.csv, irrelevant.csv. Users need to specify the number of random samples, and prediction probabilities results are saved as corresponding pickle files. The script is available at \url{https://github.com/cyberyu/ava/scripts/dropout_script.py}
	
	\subsubsection {Visualization of Model Accuracy and Uncertainty} The visualization notebook \url{https://github.com/cyberyu/ava/scripts/notebooks/BERT_dropout_visualization.ipynb} uses output pickle files from the previous script to generate histogram distribution figures and figures 4(b) and (c).
	
	\subsection{Threshold Optimization Source Code}
	\subsubsection{Threshold for entropy} Optimization script finding best threshold for entropy is available at \url{https://github.com/cyberyu/ava/blob/master/scripts/optimization/optimize_entropy_threshold.py}. The script requires Python 3.6 and Gurobi 8.1.  
	
	\subsubsection{Threshold for mean probability and standard deviation} Optimization script finding best mean probability threshold and standard deviation threshold is available at \url{https://github.com/cyberyu/ava/blob/master/scripts/optimization/optimize_dropout_thresholds.py}
	
	\subsection{Sentence Completion Source Code}
	The complete Sentence Completion RESTFUL API code is in \url{https://github.com/cyberyu/ava/scripts/sentence_completion/serve.py}. The model depends on BertForMaskedLM function from Transformer package (ver 2.1.1) to generate token probabilities. We use transformers-cli (\url{https://huggingface.co/transformers/converting_tensorflow_models.html}) to convert our early pretrained embeddings to PyTorch formats. 
	The input parameters for API are:
	\begin{itemize}
	\item{Input sentence. The usage can be three cases:
	    \begin{itemize}
        \item The input sentence can be noisy (containing misspelled words) that require auto-correction. As shown in the example, the input sentence has some misspelled words.
        \item Alternatively, it can also be a masked sentence, in the form of “Does it require [MASK] signature for IRA signup”.  [MASK] indicates the word needs to be predicted. In this case, the predicted words will not be matched back to input words.  Every MASKED word will have a separate output of top M predict words.  But the main output of the completed sentence is still one (because it can be combined with misspelled words and cause a large search) .
        \item Alternatively, the sentence can be a complete sentence, which only needs to be evaluated only for Perplexity score.  Notice the score is for the entire sentence.  The lower the score, the more usual the sentence is.
        \end{itemize}}
        \item {Beamsize: This determines how many alternative choices the model needs to explore to complete the sentence. We have three versions of functions, predict\_oov\_v1, predict\_oov\_v2 and predict\_oov\_v3. When there are multiple [MASK] signs in a sentence, and beamsize is larger than 100, v3 function is used as independent correction of multiple OOVs. If beamsize is smaller than 100, v2 is used as joint-probability based correction.  If a sentence has only one [MASK] sign, v1 (Algorithm 2 in Appendix) is used.}
        
        \item {Customized Vocabulary: The default vocabulary is the encoding vocabulary when the bidirectional language model was trained.  Any words in the sentence that do not occur in vocabulary will be treated as OOV, and will be predicted and matched.   If you want to avoid predicting unwanted words, you can include them in the customized vocabulary.  For multiple words, combine them with “|” and the algorithm will split them into list. It is possible to turn off this customized vocabulary during runtime, which simply just put None in the parameters.

        \item {Ignore rule: Sometimes we expect the model to ignore a range of words belonging to specific patterns, for example, all words that are capitalized, all words that start with numbers.   They can be specified as ignore rules using regular expressions to skip processing them as OOV words.  For example,  expression "[A-Z]+" tells the model to ignore all uppercase words, so it will not treat `IRA' as an OOV even it is not in the embeddings vocabulary (because the embeddings are lowercased).  To turn this function off, use None as the parameter.}
 }
 \end{itemize}
 
 The model returns two values: the completed sentence, and its perplexity score.	

	\subsection{RASA Server Source Code}
	
    The proposed chatbot utilizes RASA's open framework to integrate RASA's ``chitchat'' capability with our proposed customized task-oriented models. To achieve this, we set up an additional action endpoint server to handle dialogues that trigger customized actions (sentence completion+intent classification), which is specified in actions.py file. Dialogue management is handled by RASA's Core dialogue management models, where training data is specified in stories.md file. So, in RASA dialogue\_model.py file run\_core function, the agent loads two components: \textrm{nlu\_interpreter} and \textrm{action\_endpoint}.
	
    The entire RASA project for chatbot is shared under \url{https://github.com/cyberyu/ava/bot}. Please follow the github guidance in README file to setup the backend process. 
	
	\subsection{Microsoft Teams Setup}
    Our chatbot uses Microsoft Teams as front-end to connect to RASA backend.  We realize setting up MS Teams smoothly is a non-trivial task, especially in enterprise controlled enviornment.  So we shared detailed steps on Github repo.  
	
	\subsection{Connect MS Teams to RASA}
	At RASA side, the main tweak to allow MS Team connection is at dialogue\_model.py file. The BotFrameworkInput library needs to be imported, and the correct app\_id and app\_password specified in MS Teams setup should be assigned to initialize RASA InputChannel.

\end{document}